\documentclass[11pt,a4paper]{article}
\usepackage[T1]{fontenc}
\usepackage{amsmath}
\usepackage[polish]{babel}
\usepackage[utf8]{inputenc}
\usepackage[hyperref]{naaclhlt2019}
\usepackage{times}
\usepackage{latexsym}
\usepackage{graphicx}
\usepackage{cuted}
\usepackage{subfig}

\usepackage{url}

\defcitealias{ecommendation2001method}{ITUR Recommendation, 2001}
\aclfinalcopy 
\def\aclpaperid{4054} 

\title{In Other News: A \textit{Bi-style} Text-to-speech Model\\ for Synthesizing Newscaster Voice with Limited Data}

\author{Nishant Prateek,
 Mateusz \L{}ajszczak, 
 Roberto Barra-Chicote,
 Thomas Drugman, \vspace{1mm}\\
 \textbf{Jaime Lorenzo-Trueba,
 Thomas Merritt,
 Srikanth Ronanki,
 Trevor Wood} \vspace{2mm} \\ 
Amazon Research Cambridge, UK \vspace{1.5mm}  \\
{\tt nprateek@amazon.co.uk}}

\date{}

\begin{document}
\maketitle
\begin{abstract}
Neural text-to-speech synthesis (NTTS) models have shown significant progress in generating high-quality speech, however they require a large quantity of training data. This makes creating models for multiple styles expensive and time-consuming. In this paper different styles of speech are analysed based on prosodic variations, from this a model is proposed to synthesise speech in the style of a newscaster, with just a few hours of supplementary data. We pose the problem of synthesising in a target style using limited data as that of creating a bi-style model that can synthesise both neutral-style and newscaster-style speech via a one-hot vector which factorises the two styles. We also propose conditioning the model on contextual word embeddings, and extensively evaluate it against neutral NTTS, and neutral concatenative-based synthesis. This model closes the gap in perceived style-appropriateness between natural recordings for newscaster-style of speech, and neutral speech synthesis by approximately two-thirds.
\end{abstract}

\section{Introduction}
Newscasters have a clearly identifiable dynamic style of speech. As more people are using virtual assistants, in their mobile devices and home appliances, for listening to daily news, synthesising newscaster-style of speech becomes commercially relevant. A newscaster-style of speech gives users a better experience when listening to news as compared to news generated in the neutral-style speech, which is typically used in text-to-speech synthesis. In addition, synthesising news using text-to-speech is more cost-effective and flexible than having to record new snippets of news with professional newscasters every time a new story breaks in.

Recent advances in neural text-to-speech (NTTS) synthesis ~\cite{van2016wavenet, wang2017tacotron, shen2018natural, merritt2018comprehensive} have enabled researchers to generate high-quality speech with a wide range of prosodic variations. For many years, concatenative-based speech synthesis ~\cite{black1995optimising, taylor2006target, qian2013unified, merritt2016deep, wan2017google} has been the industry standard. Concatenative-based speech synthesis methods can produce high-quality speech, but are limited by the coverage of units in its database. When it comes to more expressive styles of speech, this problem is aggravated by the many hours of speech data that would be needed to cover an acceptable range of prosodic variations present in a particular style of speech. The concatenative approaches also require extensive hand-crafting of relevant low-level features, and arduous engineering efforts. 

Recently proposed models based on sequence-to-sequence (seq2seq) architecture ~\cite{wang2017tacotron,shen2018natural,ping2017deep} attempt to alleviate some of these issues by transforming the low-level feature representation into a learning task. These models function as acoustic models which take text, in the form of characters or phonemes as input, and output low-level acoustic features that can be then converted into speech waveform using one of the several `vocoding' techniques ~\cite{perraudin2013fast, shen2018natural, lorenzo2018robust}. Seq2seq models also allow us to condition the model on additional observed or latent attributes that help in improving the flexibility (modelling different speaker, and styles), and naturalness ~\cite{ping2017deep, jia2018transfer, wang2018style, skerry2018towards, stanton2018predicting}. Li et. al.~\shortcite{li2018close} have explored transformer networks for context generation. This improves training efficiency while capturing long-range dependencies. Even though transformers have enabled parallel training, they still suffer from slow inference due to autoregression. LSTM-based seq2seq architectures, having lesser number of trainable parameters, allow for faster inference. 

Several works have explored the ``controllability''  of style in synthesised speech through latent-variable modelling techniques ~\cite{akuzawa2018expressive, Henter2018DeepEM, hsu2018hierarchical}. These models not only enable us to jointly model different styles, but also allow the user to control the style through modification of disentangled latent variable during the inference. Although flexible, these models usually require a large amount of data to capture the idiosyncrasies of speaking styles, and to disentangle the characteristics of speech (pitch, duration, amplitude etc.) Additionally, these models are slow to train and are potentially overly complex for modelling styles of speech that are expressive but do not display large prosodic variations. During inference, the user would need to input the latent variables to synthesise, which is not ideal for production systems.

Conventional seq2seq models for NTTS rely on a  single encoder for linguistic inputs (phonemes/character embeddings). This encoder cannot be solely relied upon to capture higher-level text characteristics like syntax or semantics. The relation between syntax, semantics and prosody is complex. Many linguistic theories try to tie these phenomena but they struggle to explain some edge cases and are mutually inconsistent ~\cite{taylor2009text}
. Thus, it might be unsatisfactory to apply linguistic knowledge directly to prosody modelling by conditioning the model on manually selected features.  Recent advances in representation learning for text ~\cite{Peters:2018, devlin2018bert} have allowed us to come up with linguistic representations that not only capture the semantics of a word, but are also context-dependent as a function of the entire sentence. Contextual word embeddings (CWE) can be used to present to the model additional conditioning features that can help model the prosodic variations in each word, based on the context in which it is present.

Latorre et. al ~\shortcite{latorre2018effect} investigated the effect of data reduction on seq2seq acoustic models. They train a multispeaker model with limited data from several speakers. Chung et. al ~\shortcite{chung2018semi} pre-train the decoder of their acoustic model on a large amount of unpaired data where the decoder learns the task of predicting the next frame. They also propose conditioning the model on traditional word-vectors like GloVe and Word2vec~\cite{pennington2014glove, mikolov2013distributed} to provide additional linguistic information. Both these works don't look at varying prosody or speaking-style. There has been a growing interest in adaptive techniques for voice cloning ~\cite{arik2018neural, chen2018sample}, and style adaptation ~\cite{bollepalli2018speaking} with limited data. However, these models require extensive fine-tuning. Additional investigation is needed on the performance of such adaptive models on more multi-style setting.

The contribution of this work is two-fold: (1) We propose a \textit{`bi-style'} model that is capable of generating both a distinct newscaster style of speech, and neutral style of speech, trained only on few hours of supplementary newscaster-style data, (2) we explore the use of CWE as an additional conditioning input for prosody modelling.

\begin{figure*}[t!]
    \centering
    \includegraphics[width=14cm, height=8.25cm]{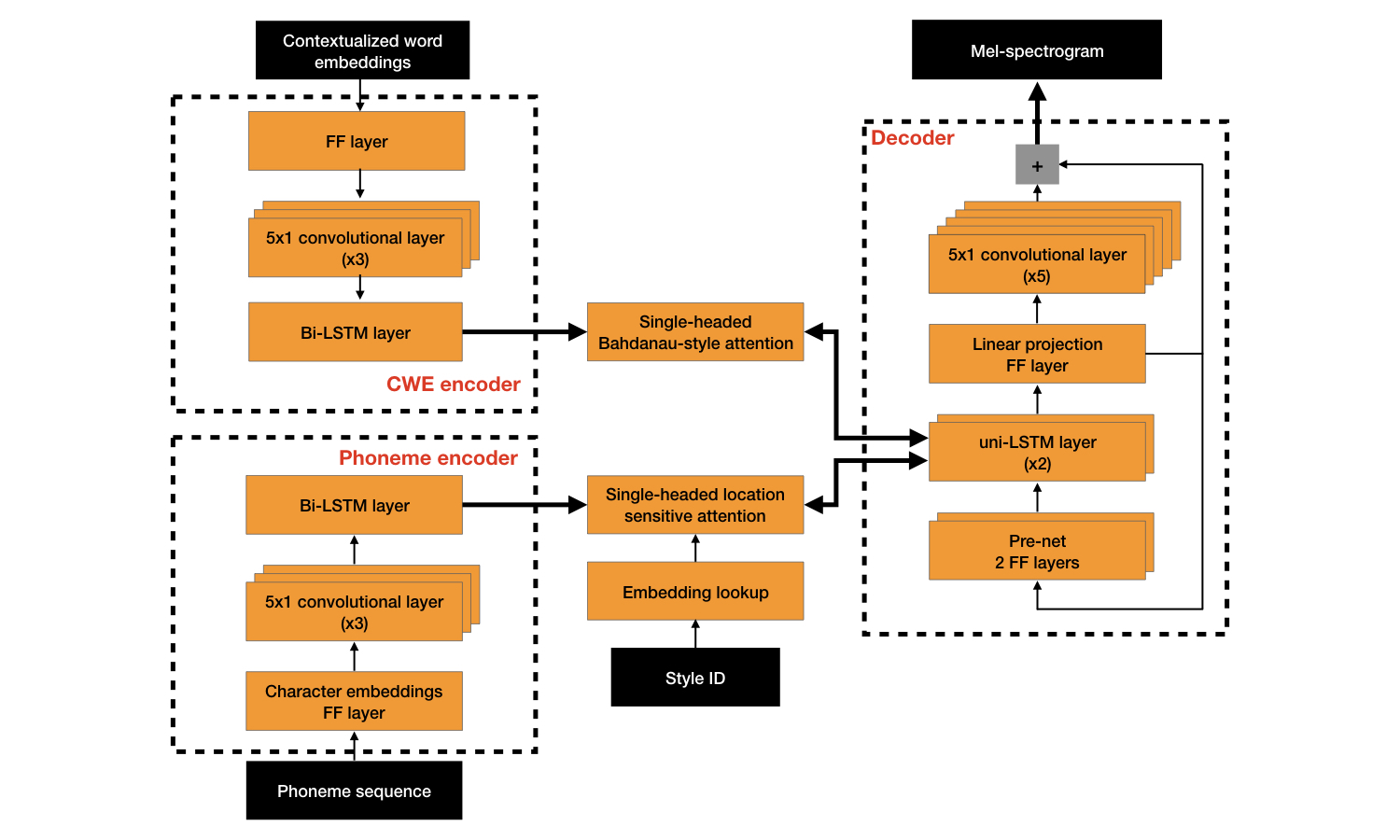}
    \caption{\label{fig:model}Context Generation Module}
\end{figure*}

\section{Data Exploration}
This section aims at understanding the prosodic variability in neutral-style, and newscaster-style corpora. For this purpose, we study the average variance in the natural logarithm of fundamental frequency (\textit{lf0}) for each utterance in the two styles. The values are reported in Table \ref{tab:prosody}. For contrast, we also study per-utterance \textit{lf0} in a mixed-expressive corpus from the same speaker. We notice that among the three corpora, the neutral-style utterances have the lowest mean variance per utterance, making it more tractable and easier to model with NTTS than the other two corpora. Newscaster-style has a slightly higher mean variance given greater expressiveness, and the mixed-expressive corpus has the highest mean variance. Latent-variable models ~\cite{akuzawa2018expressive, hsu2018hierarchical, wang2018style, Henter2018DeepEM, stanton2018predicting} tackle the problem of modelling varied expressive corpora. As we have already discussed, these models are slow to train, and require prediction or manual injection of continuous latent variables during inference. These might not be well-suited for the task of modelling newscaster-style, which even though is expressive, has much lower mean variance per utterance than the mixed-expressive corpus. 

\begin{table}[h!]
\begin{center}
\begin{tabular}{|l|c|r|}
\hline \bf Corpus & \bf Variance & \bf Range\\ \hline
Neutral & 6.32 & 5.66\\
Newscaster & 6.33 & 5.68\\
Mixed expressive & 6.79 & 5.71\\
\hline
\end{tabular}
\end{center}
\caption{\label{tab:prosody} Analysis of  mean prosodic variations based on \textit{lf0} per utterance}
\end{table}

Latorre et. al. ~\shortcite{latorre2018effect} found that a minimum of $\sim$ \(15000\) utterances (approximately \(15\) hours of data) are required to train a seq2seq acoustic model from scratch. Gathering \(15\) hours of data for each new style is both expensive and time-consuming. Given that the mean variance for the newscaster-style utterances is marginally higher than that of neutral-style utterances, we propose jointly modelling both the neutral-style and the newscaster-style, with a one-hot \textit{`style ID'} to differentiate between the two styles. We hypothesise that the style ID will be able to effectively factorise the neutral and newscaster styles, and generate style-appropriate samples for both. This will also alleviate the problem of prediction, and injection of continuous latent variables, that might introduce additional latency in the system. During inference, the style ID can be set by modification of simple binary flags.

From our internal corpus of female US-English voice, we use $\sim$ \(20\) hours of neutral-style utterances. For the newscaster-style, we use additional recordings from the same voice talent, approximating the style of American newscasters. For experiments in this paper, the amount of data used for the newscaster-style is one-fifth that of neutral-style. Using both these utterances to train a bi-style model provides us with enough overall data to train the acoustic model, and also help the model learn to factorise the two styles with the style ID input.

\section{Model Description}

Our proposed model is composed of two modules - Context Generation and Waveform Synthesis. The context generation module takes phonemes as inputs, and predicts temporal acoustic features, e.g. mel-spectrograms. The predicted acoustic features are then converted to time-domain audio waveforms by the Waveform Synthesis module. We provide additional inputs to the context generation module, in the form of `style ID' and contextual word embeddings, for better prosody modelling.

\subsection{Context Generation}
The context generation module is an extension of the seq2seq-based acoustic model proposed by Latorre et al. ~\shortcite{latorre2018effect}, and is shown in Figure \ref{fig:model}. We propose multi-scale encoder conditioning, with the acoustic model processing phoneme-level inputs, and an additional CWE encoder that processes word-level inputs.

\subsubsection{Acoustic Model}
The acoustic model consists of the \textit{phoneme encoder}, style ID input, a single-headed location-sensitive attention block, and the decoder module.  
The style ID is a two-dimensional one-hot vector (representing whether the input utterance belongs is in the neutral-style or newscaster-style), which is projected into continuous space by an embedding lookup layer to produce a \textit{style embedding}. The style embedding is concatenated at each step of the output of the phoneme encoder. Single-headed location-sensitive attention ~\cite{chorowski2015attention} is applied to the concatenated outputs. A unidirectional LSTM-layer takes the concatenated vector of the output vector of the attention block and the pre-net layer as an input.  The decoder, in each step, predicts blocks of 5 frames of 80-dimensional mel-spectrograms. We define a frame as a \(50ms\) sequence, with an overlap of \(12.5ms\). The last frame of the previous outputs is passed to the pre-net layer as input for generating the next set of frames.

\subsubsection{CWE Encoder}
We use \textit{Embeddings from Language Models} (ELMo), introduced by Peters et al. ~\shortcite{Peters:2018} for obtaining the contextual word embeddings for the input utterance. ELMo takes advantage of unsupervised language modelling task to learn rich text representations on a large text corpus. These representations can then be transferred to downstream tasks that often require explicit labels. ELMo embeddings bring a significant improvement for a variety of Natural Language Processing (NLP) tasks. They are able to capture both semantic and syntactic relations between words ~\cite{perone2018evaluation}. As such, they seem to be a good fit for modelling prosody.

\vspace{1.5mm}

For each sentence in the training set we extract ELMo features using publicly available CLI tool ~\cite{Gardner2017AllenNLP}. This model is pre-trained on the 1 Billion Word Benchmark dataset ~\cite{chelbaorsten}. We only use hidden states from the top layer of bi-directional Language Model (biLM). This produces a sequence of 1024-dimensional vectors, one for each word in a sentence. During training these vectors are fed to \textit{CWE encoder}. CWE encoder has a similar topology to the phoneme encoder. 

Encoded ELMo embeddings are passed to the decoder through Bahdanau-style attention ~\cite{bahdanau2014neural}. It operates independently of location sensitive attention for phoneme encodings. It can attend to encodings of words that are not focused by location sensitive attention. We hypothesise that this can help the decoder to consider a broader context. 

\subsection{Waveform Synthesis}
We use the pre-trained speaker-independent RNN-based ``neural vocoder'' proposed by Lorenzo-Trueba et. al. ~\shortcite{lorenzo2018robust} to convert the mel-spectrograms predicted by our context generation module into high-fidelity audio waveforms.

\section{Experimental Protocol}

\subsection{Training}
The news stories are on an average longer than neutral-style utterances, and consist of multiple sentences. Seq2seq models have a tendency to lose attention and have misalignment in longer input sequences during inference. To alleviate this, we split the news stories into individual sentences in both the training and the test sets. Splitting into individual sentences also enables us to train the model on larger batch size, helping the model to converge faster and with lesser perturbation of the training loss. To convert the utterances into phoneme sequences, we use our internal grapheme-to-phone mapping tool, which encodes the phonemes, stress marks, and punctuations as one-hot vectors.

We train the model using an L1 loss in the decoder output for mel-spectrogram prediction. To indicate when to stop predicting the decoder outputs, we have a linear stop token generator at the decoder outputs, trained jointly with the context generation module. The stop token generator is trained with an L2 loss. During training, the stop token is linearly increased from 0 at the beginning of the sentence to 1 at the end.

ADAM optimizer ~\cite{kingma2014adam} is used to minimise the training loss, with learning rate decay. The model is trained with teacher-forcing on the decoder outputs. The attention weights are normalised to add up to \(1\) using a softmax layer.

We use mel-spectrogram distortion ~\cite{kubichek1993mel} to monitor the input-output alignment, and the training loss to get a rough estimation on the convergence of our model. We also synthesise some held-out sentences to monitor the segmental quality and the prosody of our system, as the perceptual quality of the generated samples does not always align with the lower training and validation losses, and spectrogram distortion metrics. 

\begin{table*}[h!]
\begin{center}
\begin{tabular}{|l|l|}
\hline \bf System & \bf Description \\ \hline
Concatenative & Concatenative-based unit selection system driven by state-level statistical\\ & parametric predictions\\
Neutral & Neutral-style NTTS speech\\
News w/o CWE & Newscaster-style NTTS speech without CWE conditioning\\
News with CWE & Newscaster-style NTTS speech with CWE conditioning\\
Recordings & Natural speech waveforms\\
\hline
\end{tabular}
\end{center}
\caption{\label{tab:MUSHRA} Systems present in the MUSHRA evaluation}
\end{table*}

\subsection{Setup for Evaluation}
\subsubsection{Objective Metrics}
We compare acoustic parameters extracted from the synthesised sentences, and the natural recordings for the analysis of prosody and segmental quality. To match the predicted sequence length to the reference sequence length for all comparisons, we use the dynamic time warping (DTW) algorithm ~\cite{bellman1959adaptive}.  \\
We use Mel-spectrogram Distortion to assess the segmental quality of the synthesised sentences. \\ 
\noindent \textbf{Mel-spectrogram distortion (MSD)} ~\cite{kubichek1993mel} measures the distortion between predicted and extracted (from natural speech) mel-spectrogram coefficients and is defined as: \\
\begin{equation}
\label{eq:mcd}
MSD = \frac{\alpha}{T} \sum_{t=1}^{T}\sqrt{\sum_{d=1}^{D-1}(c_d(t)-\hat{c}_d(t))^2}
\end{equation}
\begin{equation}
\alpha = \frac{10\sqrt{2}}{ln10}
\end{equation}
\noindent where $c_d(t)$, $\hat{c}_d(t)$ are the d-th mel-spectrogram coefficient of the t-th frame from reference and predicted. T denotes the total number of frames in each utterance and D is the dimensionality of the mel-spectrogram coefficients. For our experiments, we use 80 coefficients per speech frame. The zeroth coefficient (overall energy) is excluded from MSD computation, as shown in equation \ref{eq:mcd}. \vspace{2mm}

For evaluating prosody, we use the following metrics calculated on \textit{lf0}:\\ 
\noindent \textbf{F0 Root Mean Square Error (FRMSE)} is defined as:
\begin{equation}
 FRMSE = \\
 \sqrt{\frac{\sum_{t=1}^{T} (x_t-\hat{x_t})^2}{T}}
\end{equation}
\noindent where ${x_t}$ and ${\hat{x_t}}$ in our work denote \textit{lf0} extracted from reference and predicted audio respectively. \vspace{2mm} \\
\noindent \textbf{F0 Linear Correlation Coefficient (FCORR)} is the measure of the direct linear relationship between the predicted \textit{lf0} and the reference \textit{lf0}. It is expressed as:
\begin{equation}
\frac{T \sum{(x_t\hat{x_t})} - (\sum{x_t})(\sum{\hat{x_t}})}{\sqrt{T (\sum{x_t^2}) - (\sum{x_t})^2}\sqrt{T (\sum{\hat{x_t}^2}) - (\sum{\hat{x_t}})^2}}
\end{equation}
\noindent If $x_t$ and $\hat{x_t}$ have a strong positive linear correlation, FCORR is close to +1. \vspace{2mm} \\
\noindent \textbf{Gross pitch error (GPE)} ~\cite{nakatani2008method} is measured as percentage of voiced frames whose relative \textit{lf0} error is more than 20\%. Relative \textit{lf0} error is defined as:
 \begin{equation}
\frac{\left|x_t-\hat{x_t}\right|}{{x_t}} \times 100
\end{equation} \\
\noindent \textbf{Fine pitch error (FPE)} ~\cite{krubsack1991autocorrelation} is measured as standard deviation of the distribution of relative  \textit{lf0} errors, for which relative \textit{lf0} error is less than 20\%. \vspace{0.2cm}

Since we don't explicitly predict \textit{lf0}, we use \textit{lf0} extracted from natural recordings, and synthesised sentences for computation of the objective metrics described above.

\subsubsection{Subjective Evaluations}
Even though the objective metrics give us a general indication on the prosody and segmental quality of synthesised speech, the metrics may not directly correlate to the perceptual quality. We conduct additional subjective evaluations with human listeners and consider these as the final outcome of our experiments.

For subjective evaluations, we concatenate the synthesised news-style sentences into full news stories, to capture the overall experience of our intended use-case. Each utterance is 3-5 sentences long, and the average duration is $33.47 seconds$. We test our system with 10 expert listeners with native linguistic proficiency in English, using the MUltiple Stimuli with Hidden Reference and Anchor (MUSHRA) methodology ~\cite{recommendation2001method}. The systems used in this evaluation are described in Table \ref{tab:MUSHRA}. The listeners are asked to rate the appropriateness of each system as a newscaster voice on a scale of \(0\) to \(100\).  For each utterance, \(5\) stimuli are presented to the listeners side-by-side on the same screen, representing the \(5\) test systems in a random order. Each listener rates \(51\) screens.

\section{Results}
\subsection{Analysis of Objective Metrics}

\begin{table*}[t!]
\begin{center}
\begin{tabular}{|c|c|c|c|c|c|}
\cline{2-6}
\multicolumn{1}{c|}{} & \multicolumn{1}{ |c| }{\textbf{Segmental Quality}} &\multicolumn{4}{|c|}{\textbf{Prosody}}\\
\hline \bf System & \bf MSD (dB) & \bf FRMSE (Hz) & \bf FCORR & \bf GPE (\%) & \bf FPE (cents)\\ \hline
Concatenative & 6.07 & 44.85 &  0.28 & 33.58 & 5.68\\
Neutral & 5.27 & 44.81 & 0.30 & 32.02 & 5.63\\
News w/o CWE & \textbf{4.52} & 42.90 & 0.35 & 28.89 & 5.57\\
News with CWE & 4.54 & \textbf{42.14} & \textbf{0.36} & \textbf{27.59} & \textbf{5.55}\\
\hline
\end{tabular}
\end{center}
\caption{\label{tab:obj_stats} Objective metrics for analysis of prosody and segmental quality. High FCORR indicates better prosody. For all other metrics, lower value indicates better performance. }
\end{table*} 

The scores for the objective metrics are shown in Table \ref{tab:obj_stats}. We observe that both of our newscaster-style models obtain consistently better scores on all metrics, than neutral NTTS and concatenative-based system. Furthermore, we also observe that conditioning the newscaster-style model with CWE helps improve the prosody of the synthesised utterances. \\
There's a slight loss in segmental quality when conditioning the model with CWE, but it appears to be imperceptible to human listeners. 
\subsection{Analysis of MUSHRA Scores}
\begin{figure}[h!]
    \centering
    \includegraphics[width=\columnwidth]{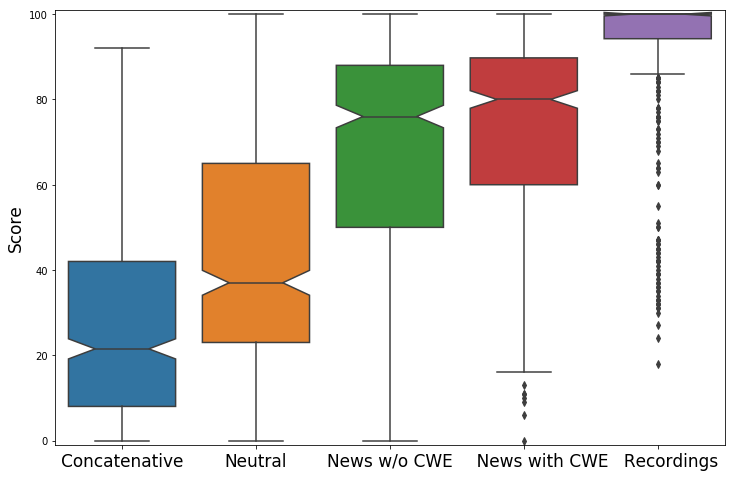}
    \caption{Boxplot of the listener responses in the MUSHRA evaluation}
    \label{fig:mushra_box}
\end{figure}

The listener responses from the subjective evaluation are shown in Figure \ref{fig:mushra_box}. In Table \ref{tab:desc_stats} the descriptive statistics for the MUSHRA evaluation are reported. 
The proposed model closes the gap between concatenative-based synthesis for newsreading, which is still largely the industry standard, and the natural recordings by 69.7\%. 
The gap compared with the neutral NTTS voice is also closed by 60.9\%.
All of the systems present in the MUSHRA test are statistically significant from each other at a p-value of 0.01. This significance is observed across the 
listener responses using a t-test. Holm-Bonferroni correction was applied due to the number of condition pairs to compare. 
This significance is also observed over the MUSHRA responses in terms of the rank order awarded by listeners. For this a Wilcoxon signed-rank test applying Holm-Bonferroni correction was used.

\begin{table*}[t!]
\begin{center}
\begin{tabular}{|c|c|c|c|c|}
\hline \bf System & \bf Mean score & \bf Median score & \bf Mean Rank & \bf Median Rank\\ \hline
Concatenative & 28.31 & 21.5 & 4.60 & 5\\
Neutral & 42.44 & 37.0 & 3.86 & 4\\
News w/o CWE & 68.15 & 76.0 & 2.67 & 3\\
News with CWE & 72.4 & 80.0 & 2.41 & 2\\
Recordings & 91.61 & 100.0 & 1.45 & 1\\
\hline
\end{tabular}
\end{center}
\caption{\label{tab:desc_stats} Listener ratings from the MUSHRA evaluation }
\end{table*}

The concatenative-based system is prone to audible artefacts at the concatenation-points, primarily due to abrupt changes in fundamental frequency in voiced phonemes. This reduces the perceived naturalness of synthesised speech. The neutral-style system is unable to model the prosody that is distinct to the newscaster-style of speech. A higher score for the newscaster-style model with CWE conditioning with respect to the model without, provides evidence supporting the hypothesis that we made in Section 1 that CWE features help model the prosodic variation better given the additional information on the syntactic context of words in the sentence.

\begin{figure}[h!]
    \centering
    \includegraphics[width=\columnwidth]{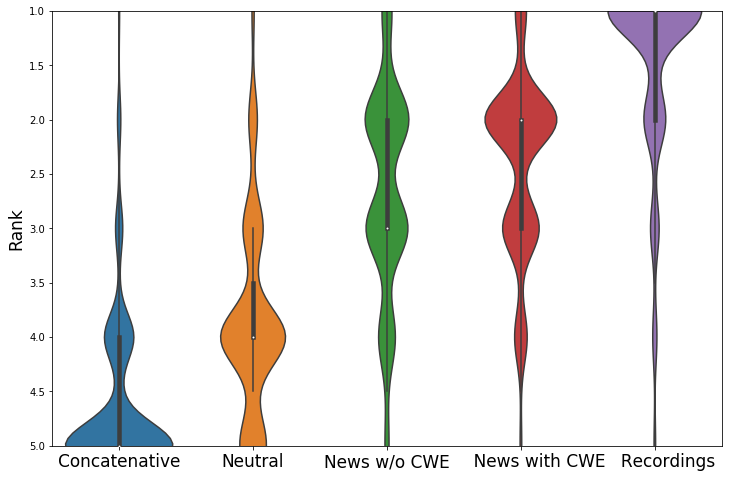}
    \caption{Violin plot of the rank-order awarded by listeners}
    \label{fig:violin}
\end{figure}

We also generated a violin plot (Figure \ref{fig:violin}) depicting the distribution of the rank-order awarded to the systems in the test. We notice that for some of the utterances, the listeners have ranked our newsreader voice (both with and without CWE) higher than the natural recordings, showing that our context generation module is able to closely mimic the recordings in terms of prosody and naturalness. 

\subsection{Effect of Contextual Word Embeddings on Prosody Modelling}
To further reinforce the effect of CWE on prosody modelling for newscaster-style, a preference test was conducted comparing newscaster-style with and without CWE conditioning, using 10 expert listeners. Listeners were informed to rate the systems in terms of their naturalness, and were asked to choose between News with CWE, News w/o CWE, or indicate \textit{No Preference}(NP).

\begin{table}[h!]
\begin{center}
\begin{tabular}{|l|r|}
\hline \bf Preference & \bf Votes \\ \hline
News with CWE & 43.2\%\\
News w/o CWE & 31\%\\
No Preference & 25.8\%\\
\hline
\end{tabular}
\end{center}
\caption{\label{tab:pref} Preference test between systems with and without CWE conditioning }
\end{table}

The listener responses are shown in Table \ref{tab:pref}. The samples conditioned on contextual word embeddings are shown to be significantly preferred (\(43.2\%\)) over the samples generated without (\(31\%\)), with \(p<0.01\). A binomial test was used to detect statistical significance.

\subsection{Analysis of Speech Tempo}
\vspace{1mm}
We define speech tempo of a corpus as the average number of phonemes present per second. Speech tempo is a crucial aspect in differentiating between the neutral and the newscaster styles. The newscaster-style is more dynamic than the neutral-style utterances, with higher speech tempo. In Table \ref{tab:tempo} we report the speech tempo in the neutral-style, and the newscaster-style for natural recordings, and compare those with our models with and without CWE. We observe that the model conditioned on CWE can better model the speech tempo in both styles. This gives us additional evidence that conditioning the model on CWE helps us synthesise samples that are not only more style-appropriate, but are also better in naturalness with respect to natural recordings. 
\begin{table}[h!]
\begin{center}
\begin{tabular}{|l|c|r|}
\hline \bf System & \bf  Neutral & \bf Newscaster\\ \hline
Recordings & 11.63 & 14.02\\
with CWE & 10.12 & 13.88\\
w/o CWE & 10.11 & 13.65\\

\hline
\end{tabular}
\end{center}
\caption{\label{tab:tempo} Speech tempo: recordings vs test systems }
\end{table}
Analysis of speech tempo also shows us that the model is able to factorise, and replicate during inference, both styles using just a one-hot style ID.

\section{Conclusions}
We proposed a bi-style model for generating neutral and newscaster styles of speech. We also proposed multi-scale encoder conditioning, focusing on phoneme-level and word-level inputs. Our proposed model is shown to be able to generate high-quality newsreader voice, which is significantly preferred over the neutral-style voice. We showed that the two styles can be factorised using a one-hot style ID. We also showed that the introduction of CWE conditioning significantly improves the prosody modelling ability of our context generation module, and hope that this result inspires more research into the use of NLP features in NTTS. 

\bibliography{naaclhlt2019}
\bibliographystyle{acl_natbib}

\end{document}